\documentclass{article}
\usepackage[a4paper,margin=1in]{geometry}
\usepackage[T1]{fontenc}
\usepackage{times}
\usepackage{microtype}
\usepackage{graphicx}
\usepackage{amsmath}
\usepackage{amssymb}
\usepackage[numbers,sort&compress]{natbib}
\usepackage[pdftex,linkcolor=blue,citecolor=blue,backref=page]{hyperref}
\usepackage{booktabs}
\usepackage{tabularx}
\usepackage{array}
\usepackage[section]{placeins}
\usepackage{afterpage}
\usepackage{capt-of}
\usepackage{eso-pic}
\AddToShipoutPictureFG*{%
  \AtPageUpperLeft{%
    \put(72,-35){\includegraphics[width=3cm]{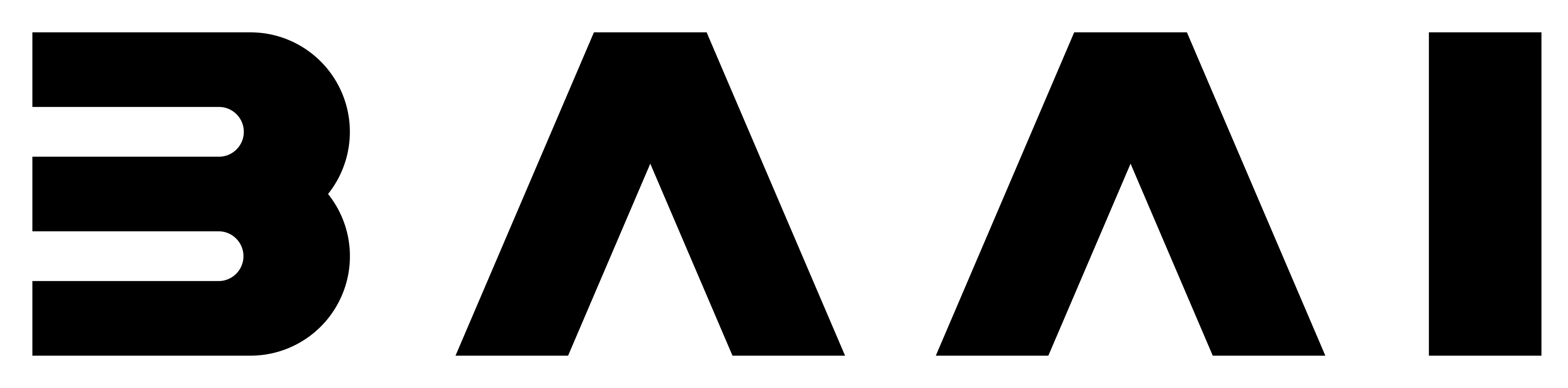}}%
    \put(72,-43){\rule{\textwidth}{0.4pt}}%
  }%
}

\newcolumntype{Y}{>{\raggedright\arraybackslash}X}
\newcolumntype{C}{>{\centering\arraybackslash}X}

\newcommand{\etal}{\emph{et al. }}

\newcommand{\eg}{\emph{e.g., }}

\newcommand{\corresponding}{\textsuperscript{\ensuremath{\dagger}}}
\newcommand{\internship}{\textsuperscript{\ensuremath{\star}}}

\renewenvironment{abstract}{%
  \section*{\centering\abstractname}%
  \normalsize\ignorespaces
}{\par}
\date{}

\title{\bfseries When the Past Shapes Action:\\Understanding Embodied Memory}
\author{%
Siqi Li\textsuperscript{1,2\internship},
Ziyi Bai\textsuperscript{3},
Xinrun Xu\textsuperscript{1,2},\\[0.3em]
Tinglei Huang\textsuperscript{1,2},
Börje F. Karlsson\textsuperscript{3}\corresponding\\[0.8em]
\small\textsuperscript{1}Institute of Software, Chinese Academy of Sciences\\
\small\textsuperscript{2}University of Chinese Academy of Sciences\\
\small\textsuperscript{3}Beijing Academy of Artificial Intelligence (BAAI)
}

\begin{document}

\maketitle
\begingroup
  \renewcommand{\thefootnote}{}
  \footnotetext{\textsuperscript{\ensuremath{\dagger}}\ Corresponding author.}
  \footnotetext{\textsuperscript{\ensuremath{\star}}\ Work performed during the author's internship at BAAI.}
\endgroup
\afterpage{%
  \clearpage
  \noindent
\begin{minipage}{\textwidth} 
    \centering
    \includegraphics[width=\linewidth,height=0.53\textheight,keepaspectratio]{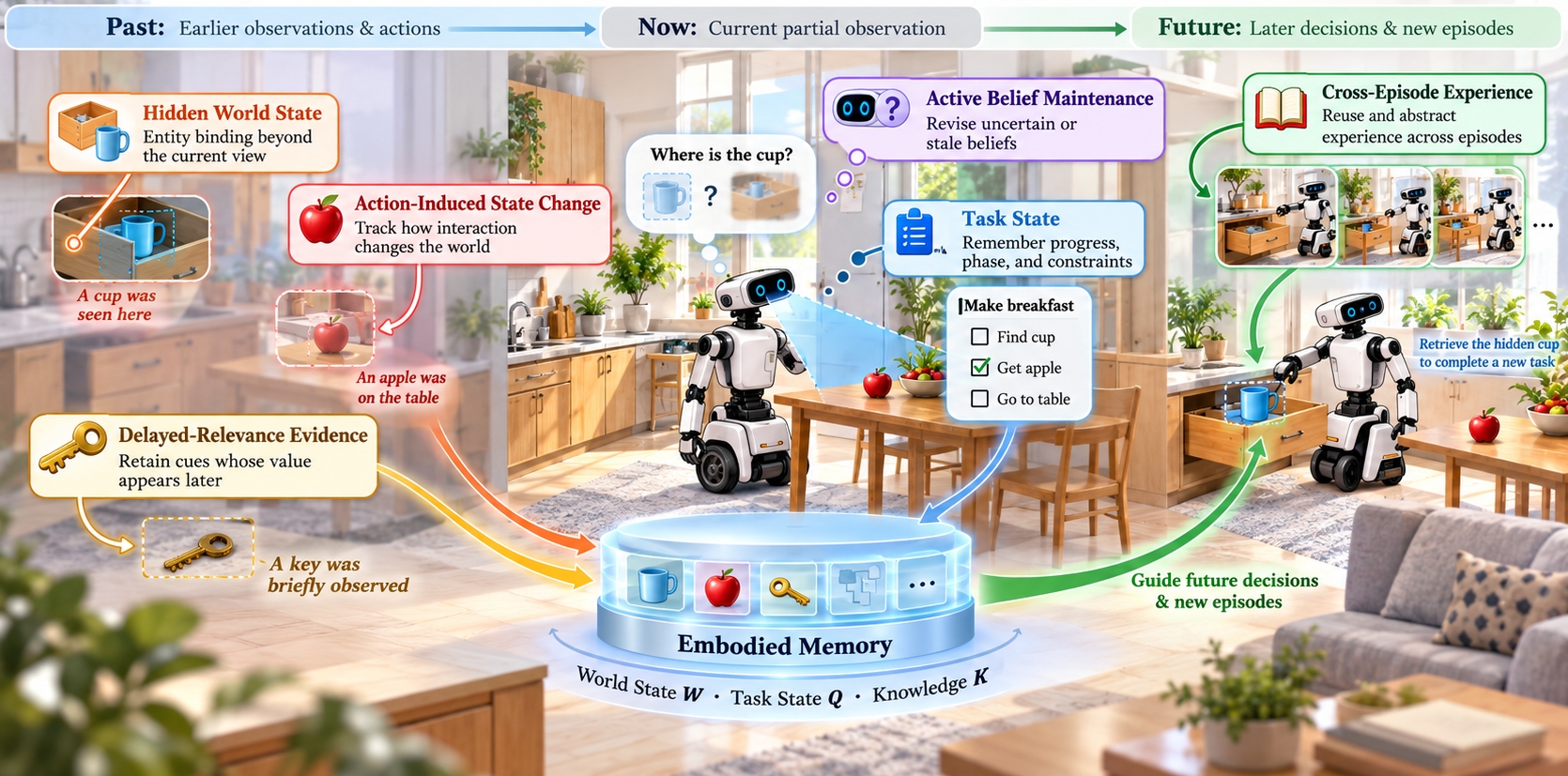}
    \captionof{figure}{\textbf{Overview of the survey.} The scene illustrates how observations and interaction outcomes become unavailable to current perception yet remain relevant to later decisions. Embodied memory preserves this evidence as world state, task state, and experience through repeated writing, updating, and retrieval.}
    \label{fig:survey_overview}
\end{minipage}
  \par\medskip
}

\begin{abstract}
Recent advances in generative foundation models have expanded the capabilities of embodied AI, enabling increasingly general navigation, manipulation, and long-horizon task execution. However, effective interaction with dynamic environments requires more than only strong perception and reasoning: agents must act with partial knowledge and retain, retrieve, and update information from past observations, actions, and experiences. This survey examines embodied memory as a mechanism for preserving decision-relevant information beyond the current observation. We organize existing approaches along two complementary dimensions: six fundamental memory requirements, spanning world-state persistence, task progress, belief maintenance, and experiential learning; and eight architectural categories, including recurrent states, fast weights, visual and symbolic representations, spatial memory, and experience-based systems. We analyze their capabilities and limitations, and examine how existing benchmarks evaluate memory-dependent behavior, distinguishing genuine use of historical information from general task performance. Finally, we identify open challenges in selecting relevant experiences, maintaining consistent beliefs in changing environments, and consolidating knowledge across episodes. Our analysis provides a framework for understanding memory as an essential component of reliable, adaptive, and lifelong-horizon embodied intelligence.
\end{abstract}

\section{Introduction}
\label{sec:introduction}

\begin{quote}
\emph{``All those moments will be lost in time, like tears in rain.''}\\
\hfill --- Roy Batty, replicant \textit{(Blade Runner, 1982)}
\end{quote}

Embodied intelligence depends on proper environment grounding and, critically, on information whose relevance outlasts its visibility. A robot may need to retrieve an object last seen inside a drawer, verify whether an earlier manipulation succeeded, or resume a procedure after an interruption. In each case, the current observation leaves decision-relevant variables unresolved, and successful action requires a state inferred from interaction history, $z_t=f(o_{1:t},a_{1:t-1})$. This challenge extends beyond recalling what a camera once observed: actions affect the world, failures reveal hidden constraints, and previous experience may alter which strategy is appropriate. Memory must therefore maintain a usable relationship between past evidence and present decisions. This perspective connects embodied memory with state estimation under partial observability and with accounts of autonomous and spatial intelligence centered on constructing and revising internal world state~\cite{lecun2022path,tos2026}.

While recent advances in generative foundation models have expanded the perceptual, reasoning, and planning capabilities of embodied agents, the underlying challenges of maintaining and exploiting historical information are longstanding. Earlier research on belief-state planning, spatial mapping, and episodic control established complementary approaches to preserving and using past experience, providing useful foundations for understanding the requirements and limitations of modern embodied memory systems (\eg\cite{Kaelbling1998,Gupta2017,Pritzel2017}).

\vspace{\baselineskip}
Recent work offers increasingly diverse mechanisms for carrying embodied history into action, but their diversity in approach and scope makes comparisons difficult. Recurrent policies compress observations into evolving latent states~\cite{rememvla,muvla}; event memories select sparse visual evidence~\cite{eventvla2026,kemo}; textual stores summarize task progress and constraints~\cite{mem,being-0,mimir}; spatial representations preserve entities across viewpoints~\cite{3dmem,dynamem,grapheqa}; and experience libraries reuse trajectories or consolidate skills~\cite{mart,physmem,practice}. These designs preserve different information and introduce different failure modes. A compact state may retain a counter while losing an object's appearance, whereas a detailed visual archive may preserve obsolete location information without recognizing that they may have been superseded. Consequently, architectural sophistication and storage capacity alone provide an incomplete explanation of when memory helps. The useful starting point is the historical variable that a future decision requires and the operations needed to keep that variable valid.

This survey organizes embodied memory through two complementary axes: six task requirements and eight architectural design categories. The requirements describe why historical information matters: hidden world state, action-induced state change, task state, delayed-relevance evidence, active belief maintenance, and cross-episode experience. The architectural categories describe how systems retain and use such history: recurrent states, fast weights, persistent visual tokens, event or keyframe stores, textual or symbolic stores, graph or 3D stores, experience or skill representations, and hybrid compositions. These are different analytical levels with a many-to-many relationship; they are not competing counts of memory types. The architectural grouping is a practical comparison framework rather than a partition into mutually exclusive substrates: event selection can operate over visual entries, experience can be stored in text, and hybrids combine several mechanisms, for example. 

Moreover, evaluation of embodied memory approaches must establish which historical information actually matters in changing behaviour. Long episodes may remain locally solvable, while a brief episode can require memory when an early cue disappears before the decisive action. General embodied benchmarks measure valuable combinations of perception, planning, and control, yet their aggregate success rates do not identify the contribution of history. Dedicated suites such as RMBench~\cite{rmbench2026}, RoboMME~\cite{robomme2026}, and MIKASA-Robo~\cite{mikasa2026} create more explicit dependencies through counting, occlusion, and delayed recall. Even there, the observation interface, supplied demonstrations, and task-progress scaffolding determine what the agent itself must remember. We therefore examine concrete tasks, the historical variables they require, the outcomes they score, and the alternative explanations that remain. Our analysis distinguishes retaining evidence from revising stale state, and successful reuse of a specific past case from the formation of transferable knowledge.

For the purposed of this survey, We define embodied memory as a latent or explicit state that preserves task-relevant evidence from earlier embodied observations, actions, outcomes, or demonstrations and makes it usable after that evidence is absent from the current observation. The scope includes navigation, manipulation, task and action planning, and spatial or egocentric reasoning about previously encountered scenes and goals, with passive video evaluation distinguished from closed-loop execution. Retained context counts when it carries historical evidence into a later decision; a fixed window supplies bounded retention, while recurrent states, external stores, and fast weights offer other persistence mechanisms. Static instructions, generic pretraining knowledge, and parameter adaptation without demonstrated historical dependence do not by themselves establish embodied memory. Generic chatbot memory, document retrieval, computer-use agents, and hardware-memory optimization remain outside the core scope of this survey. Section~\ref{sec:definition} develops these boundaries, including the conditional roles of mapping, world models, and test-time learning.

Our literature selection supports a diagnostic narrative synthesis rather than an exhaustive systematic review. The collection prioritized relevant work from recent established conferences and journal issues, while inclusion remained conditional on a substantive memory mechanism or evaluation contribution. We combine broad searches for embodied or robot memory with terms for episodic experience, spatial maps, recurrent state, retrieval, skill consolidation, fast weights, and memory evaluation, then trace references, citations, and relevant conference records. Paper contents, official proceedings, code repositories, and project pages establish the mechanism and evaluation settings. Inclusion requires either an identifiable mechanism for retaining embodied history or an evaluation that explicitly tests historical dependence. The synthesis prioritizes coverage of distinct computational problems, and its claims about architectural suitability are analytical interpretations unless directly attributed to an experiment.

The survey contributes a requirement-centered taxonomy, an analysis of architectural trade-offs, and a task-level audit of evaluation evidence. Section~\ref{sec:nonmarkovian} identifies the six requirements and explains their overlap; Section~\ref{sec:memory_architectures} relates them to the eight design categories and the bottlenecks of selection, compression, access, revision, and transfer. Section~\ref{sec:benchmarks} distinguishes what representative benchmarks score from what their results establish about memory. Section~\ref{sec:evaluation} consolidates the resulting evaluation principles and research priorities, and Section~\ref{sec:conclusion} presents some final considerations. Across these analyses, the survey organizing question centers on which information must survive, remain valid, and become available at the decision that necessitates it. Answering that question connects memory design to observable behavioural consequences and provides a firmer basis for comparing methods than the presence of a memory module per se.

\section{What Counts as Embodied Memory?}
\label{sec:definition}

The term \emph{memory} is used broadly in embodied AI, ranging from short observation histories and recurrent hidden states to explicit episodic stores, spatial maps, and fast-weight states.
These mechanisms all expose a policy to information beyond the instantaneous sensory input, yet they provide different forms of historical accessibility.
In this survey, \textbf{embodied memory is a task-relevant state, latent or explicit, that preserves information from earlier embodied observations, actions, outcomes, or demonstrations and allows it to influence a later decision when the originating evidence is absent from the current observation}.

We use a generic update-and-use formulation,
\begin{equation}
M_t=\mathcal{U}(M_{t-1},o_t,a_{t-1}),
\qquad
a_t\sim\pi(a_t\mid o_t,g,M_t),
\label{eq:memory_update}
\end{equation}
where $M_t$ may be retained historical context, a latent state, a set of retrievable entries, a spatial representation, or an adaptive parameter state.
Three properties characterize the resulting design space: \textbf{persistence}, or how long historical information remains behaviorally available; \textbf{addressability}, or whether a specific past event or state can be selectively accessed; and \textbf{update dynamics}, or how interaction changes the retained state.
This section establishes the conceptual boundary of embodied memory; method-specific architectural comparisons are deferred to Sec.~\ref{sec:memory_architectures}.

\begin{figure}
    \centering
    \includegraphics[width=1\linewidth]{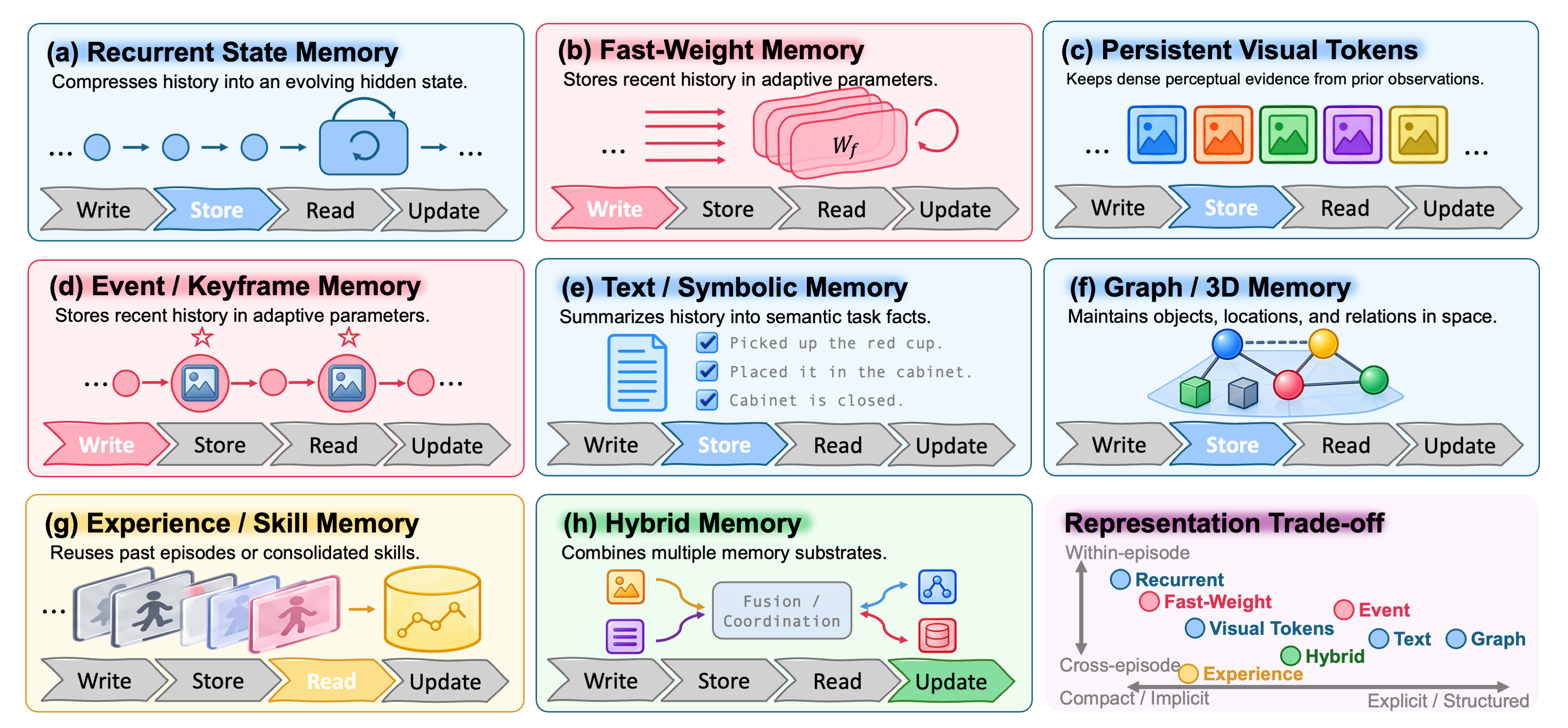}
    \caption{Conceptual boundaries and implementations of embodied memory. The six requirements in Sec.~\ref{sec:nonmarkovian} describe task demands; the architectural categories in Sec.~\ref{sec:memory_architectures} describe implementation choices. }
    \label{fig:memory_boundary}
\end{figure}

\subsection{Current-Observation Policies as No-Persistent-Memory Baselines}

A useful reference point is a policy whose decision is formed from the current observation and task specification without a dedicated persistent historical state.
Generalist robot policies such as RT-1, Diffusion Policy, ACT, OpenVLA, Octo, and $\pi_0/\pi_{0.5}$~\cite{rt1,diffusionpolicy,act,openvla,octo,pi05} provide representative baselines in this regime.
They may consume multiple camera views, exploit large pretrained representations, or generate temporally extended action chunks.
Those capabilities can produce temporally coherent behavior without preserving task-relevant evidence after it leaves the observation interface.
The relevant boundary is therefore \emph{historical accessibility}: whether information from an earlier interaction remains available when a later decision depends on it.

\subsection{Frame Stacking and Fixed-Window Temporal Context}

The most direct extension beyond instantaneous perception is a bounded observation history,
\begin{equation}
C_t^{H}=\{o_{t-H+1},\ldots,o_t\},
\label{eq:fixed_window}
\end{equation}
where $H$ defines how far the policy can directly inspect the past.
Classic frame stacking follows this principle~\cite{mnih2015human}; recent VLA systems use sparse sampling, feature aggregation, or token compression to represent several observations more efficiently~\cite{cronusvla,contextvla,zhang20254dvla}.

Modern robot policies also contain several different temporal horizons that should not be conflated.
The \emph{retrospective horizon} describes how much past observation history is accessible, the \emph{action horizon} describes how many future actions are produced in one prediction, and the \emph{predictive horizon} describes how far a model predicts future states or observations.
Action chunking and future prediction can improve control or representation learning without enlarging the history available to the policy.

The defining property of a fixed window is that retention is governed by temporal position.
An observation remains accessible while it lies inside the window and disappears when it falls outside.
This creates a structural mismatch between recency and decision relevance: redundant recent frames may remain available while an earlier decisive event has already vanished.
Fixed-window conditioning is therefore a form of \emph{bounded context memory}, intermediate between instantaneous observation and persistence beyond the window.
It belongs within our comparison, but success while a cue remains in the window does not establish retention after that cue has been evicted.

\subsection{Recurrent State and Belief-State Memory}

Recurrent memory replaces an explicit sequence of past observations with an evolving latent state,
\begin{equation}
M_t=f_\theta(M_{t-1},o_t,a_{t-1}).
\label{eq:recurrent_memory}
\end{equation}

In partially observable environments, the relevant history can be represented as a belief over latent world states, updated as observations and actions become available. Classical formulations of partially observable Markov decision processes (POMDPs) formalized this relationship between historical evidence, belief-state estimation, and decision-making under uncertainty~\cite{Kaelbling1998}. Recurrent memory can approximate such beliefs or other compact statistics of interaction history.
Early recurrent agents explored the same principle through hidden states and learned memory representations~\cite{merlin,recurrentpomdp,scenememorytransformer,episodictransformer}; recent embodied policies use recurrent queries, memory tokens, state-space representations, or latent beliefs~\cite{rememvla,muvla,rbvla,aem}.

The conceptual shift from a fixed window is the retention rule.
Fixed-window methods preserve observations because they are recent; recurrent systems preserve information because it survives repeated learned compression.
This removes an explicit temporal cutoff but introduces an information bottleneck: historical variables omitted from the evolving state may become difficult to recover later.
Recurrent memory is therefore most natural when the required history can be summarized as a compact evolving state, such as task phase, recent action effects, or a continuously updated belief.

\subsection{Persistent External Memory}

External memory keeps multiple historical entries outside the policy's transient hidden state. Representative systems use visual observations, spatial entities, textual summaries, events, trajectories, or hybrid representations (\eg~\cite{neuralmap,structuredscenememory,memonav,3dmem,memoryvla,mem,mimir}). Its defining property is \emph{addressability}: the interaction history remains partitioned into entries that can be selectively retrieved at a later decision point.

This changes the main bottleneck. A recurrent state asks whether the required history can be compressed into a compact representation; an external store asks whether the relevant historical entry can be found when it is needed. Addressability is particularly useful when a future decision depends on a specific object, event, observation, or previous episode. The concrete choices of visual, textual, spatial, episodic, or hybrid storage are analyzed in Sec.~\ref{sec:memory_architectures}.

\subsection{Alternative Update and Read Dynamics}

Historical information can also persist through mechanisms that do not resemble conventional recurrent states or explicit databases. Test-Time Training (TTT) is a useful boundary case because part of the model state changes during inference. A generic fast-weight update can be written as
\begin{equation}
W_t=W_{t-1}-\eta\nabla_W\mathcal{L}_{\mathrm{write}},
\label{eq:ttt_memory}
\end{equation}
with subsequent predictions computed using the updated $W_t$. 

TTT becomes a memory mechanism when these fast weights carry information acquired from preceding observations, demonstrations, or interactions and that information affects later decisions. When parameter updates mainly capture deployment-time distribution shift or optimize policy behavior from feedback, their primary role is adaptation. Recent systems illustrate both sides of this boundary~\cite{tttlayer,spatialttt,robott,wamttt,evolvevla,ttvla}.

In-Context Learning (ICL) provides a complementary read mechanism. Examples supplied directly with the current task can specify intended behavior, as in Zero-WAM\cite{zerowam}; this alone does not establish retention of an earlier deployment experience.

When examples are retrieved from persistent past experience, retrieval constitutes memory access and ICL becomes the interface through which selected experience influences the current decision, as illustrated by MART~\cite{mart}.

MoEC exposes a different boundary between memory and adaptation: a subgoal-indexed expert store routes Minecraft control, while new lightweight experts are trained online on a frozen backbone~\cite{moec2026}. The persistent routing store and the parameter-learning process should therefore be identified separately.

The broader lesson is that historical information may live in context, activations, fast weights, or external stores; the storage substrate alone does not determine whether the mechanism satisfies our definition of embodied memory.

\subsection{Environment Maps}

Spatial maps occupy an important boundary because they aggregate observations over time while serving different computational roles. We distinguish three levels: \textbf{geometric state estimation}, which maintains pose, occupancy, and geometry; \textbf{semantic environment memory}, which maintains persistent objects, attributes, and spatial relations; and \textbf{episodic spatiotemporal memory}, which additionally records what was observed, where it occurred, and how it changed.

Neural SLAM~\cite{neuralslam} lies near the geometric end of this spectrum, while 3D-Mem, Embodied-RAG, MemoryEQA, and RoboMemory incorporate increasingly semantic or episodic state~\cite{3dmem,embodiedrag,memoryeqa2025,robomemory}.

RANGER provides a concrete semantic memory-map example: its keyframe bank retains geometry, semantics, and exploration values from monocular observations, including an optional prior traversal video, and supports subsequent waypoint selection~\cite{ranger}. Its memory role comes from reusing scene evidence, rather than from contextual adaptation alone. MapNav offers a direct instance of semantic-map memory: its episode-level annotated map is updated over navigation and substitutes for historical image frames~\cite{mapnav2025}. This places it within the functional scope above and in A6 below.

A map has a clear memory role when historically acquired, task-relevant information remains accessible after the corresponding scene leaves sensory view and later affects action or reasoning. World models provide a related boundary: their defining function is to model how states evolve, while memory concerns which information from previous interaction remains available. A world-model state can serve both functions when it integrates historical evidence and carries that evidence into later decisions~\cite{worldmodels,planet,dreamer,memorywam}.

\section{A Taxonomy of Non-Markovian Embodied Memory Requirements}
\label{sec:nonmarkovian}

Embodied memory becomes necessary when the current observation fails to provide a sufficient state for decision making. The underlying physical process may be approximately Markovian in its complete state $s_t$, while an embodied agent receives only a partial observation $o_t$.
We call a decision \emph{history-dependent} when two situations have similar current observations and the same goal, yet different histories imply different optimal actions:
\begin{equation}
o_t^{(1)}\approx o_t^{(2)},\qquad
g^{(1)}=g^{(2)},\qquad
h_t^{(1)}\neq h_t^{(2)}
\Rightarrow
a_t^{*(1)}\neq a_t^{*(2)}.
\label{eq:history_dependence}
\end{equation}
Equivalently, some decision-relevant latent variable cannot be recovered reliably from $o_t$ alone and must be inferred from interaction history.
This formulation connects embodied memory to latent-state estimation under partial observability and to broader views of autonomous and spatial intelligence~\cite{lecun2022path,tos2026}.

We characterize each requirement through three questions: \emph{what latent state must remain recoverable}, \emph{why the current observation is insufficient}, and \emph{what computation is required to maintain the missing state}.
The latent state spans three broad forms: world state $W$, task state $Q$, and persistent experience or knowledge $K$.
These states evolve differently: world state changes with observation and physical intervention, task state changes with procedural progress and failure, and long-term knowledge consolidates across repeated experience.
The six requirements below describe recurring problem structures that follow from these different forms of historical dependence.

\textbf{These are six requirements, not six memory modules.} We use the R1--R6 identifiers in the order below. They form overlapping diagnostic dimensions: R1--R3 concern missing world or task state, R4 concerns selection before future use, R5 concerns uncertainty and evidence acquisition, and R6 concerns reuse across episodes. A task may therefore instantiate several requirements simultaneously. For example, retrieving a previously moved object requires entity persistence (R1), supersession of its old location (R2), and possibly re-observation when the move is uncertain (R5). The three state domains $W,Q,K$ are organizational aids, not additional memory types. Section~\ref{sec:memory_architectures} introduces a separate architectural axis; neither its category count nor a benchmark's own terminology changes the six requirements defined here.

\begin{figure}
    \centering
    \includegraphics[width=1\linewidth]{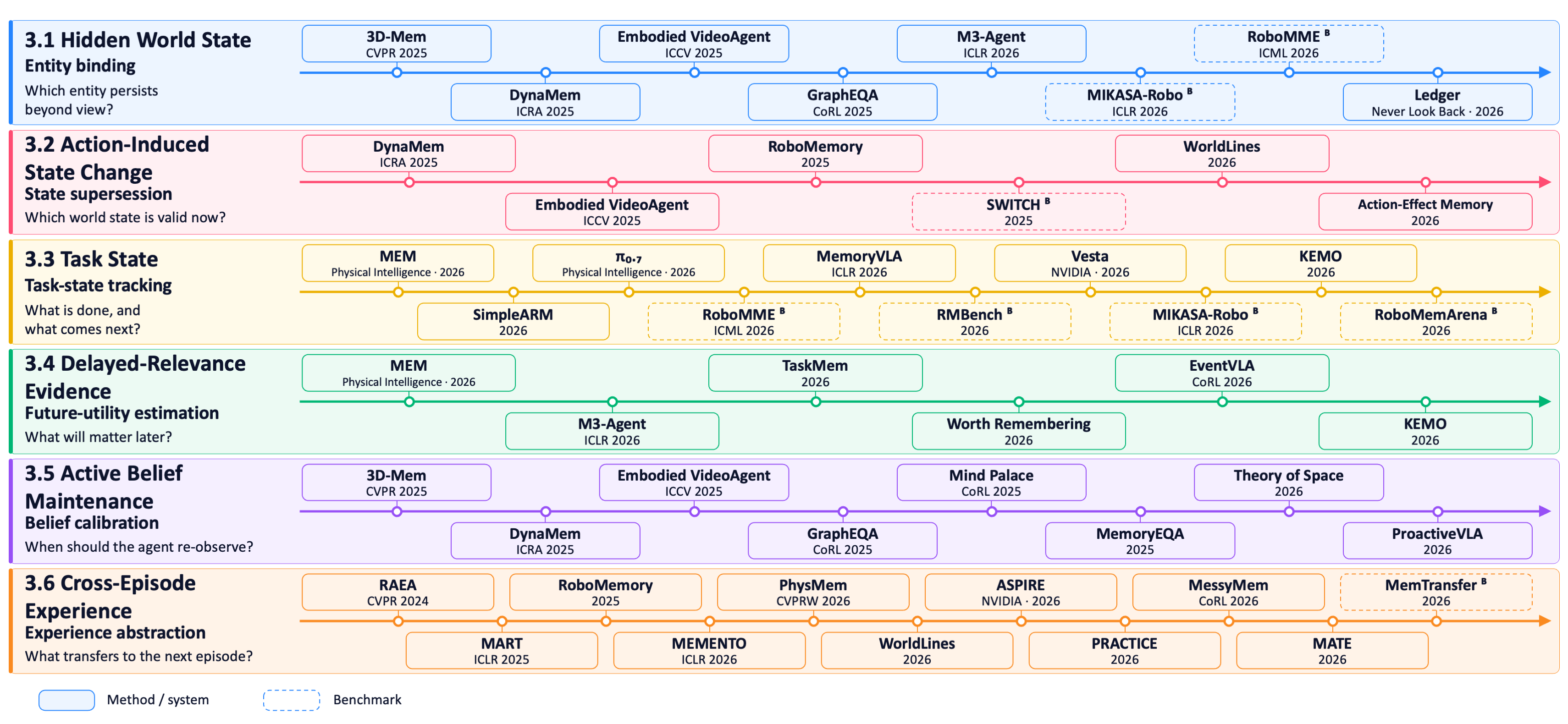}
    \caption{\textbf{Representative works organized by embodied memory requirement.} Six timelines group studies by the historical information an agent must preserve: hidden world state, action-induced state change, task state, delayed-relevance evidence, active belief maintenance, and cross-episode experience. Each row identifies the corresponding computational challenge and arranges works by year. Solid boxes denote methods or systems; dashed boxes marked B denote benchmarks. Works may appear in multiple rows when they address several requirements. Timelines are independently spaced for readability, with no ordering implied among works from the same year. Dates follow the cited bibliography, using the conference year where available and the preprint or report year otherwise.}
    \label{fig:requirements_timeline}
\end{figure}

\subsection{Hidden World State: Persistence and Entity Binding - R1}

\textbf{A diagnostic scenario:} A robot places a cup into a drawer, closes it, and five minutes later is asked to ``bring me the cup''. The current visual observation contains only the closed drawer. The robot must preserve the cup's identity and location across occlusion and viewpoint change.

The physical world extends beyond the current field of view.
Objects remain present after becoming occluded, entering containers, or leaving the camera view, so a useful memory must preserve persistent representations of entity identity, location, state, and spatial relations. The core computational problem is \emph{temporal entity binding}: associating observations acquired at different times and viewpoints with the same persistent entity. This requirement includes object permanence, re-identification, container membership, and spatial persistence.

Representative work now spans explicit 3D scene memory, dynamic spatial memory, persistent object memory from egocentric sensing, and entity-centric multimodal long-term memory. 3D-Mem, DynaMem, Embodied VideoAgent, GraphEQA, and M3-Agent~\cite{3dmem,dynamem,embodiedvideoagent,grapheqa,m3agent} provide complementary examples of these directions, while RoboMME~\cite{robomme2026}expose controlled memory-dependent variants for robotic.

LEDGER~\cite{ledger2026} further preserves object locations, movement histories, and contextual descriptions after objects leave an egocentric view. Its offline video-based evaluation tests spatial recall and localization; it does not by itself establish closed-loop control benefits.

And MapNav~\cite{mapnav2025} showcases how historical spatial evidence can also be compressed into an annotated semantic map used directly by a navigation VLM. Its value for R1 concerns maintaining navigationally useful scene state; though this alone does not establish preservation of exact visual appearance or complete object histories.

Current approaches typically rely on similarity-based persistence or explicit association. Similarity retrieval can confuse visually similar instances; explicit spatial association depends on pose and data association that can accumulate error during occlusion and motion. Binding failures appear as duplicated entities, identity swaps, or forgotten locations. Long-occlusion re-identification together with reliable incremental addition of new objects remains difficult.

\subsection{Action-Induced State Change: From Persistence to Versioned Belief - R2}

\textbf{A diagnostic scenario:} A robot sees an apple on a table, picks it up, and is then asked where the apple is. A memory that preserves only the historical fact ``apple on table'' gives the wrong answer because the agent's own action has changed the world.

Embodied agents continuously intervene in their environment. Picking up an object, opening a drawer, pressing a switch, or placing an item into a receptacle changes the state relevant to future behavior.

Memory must therefore preserve history while distinguishing historical state from the agent's current belief about the world. The core computational problem being \emph{state supersession}: newer evidence and action consequences must determine which state is currently valid without erasing useful episodic history.

DynaMem explicitly updates a spatio-semantic memory as objects move, appear, or disappear; Embodied VideoAgent couples persistent object memory with VLM-based updates when object-related actions are detected. WorldLines, Action-Effect Memory, and RoboMemory further expose state versioning, action--effect tracking, and lifelong state maintenance~\cite{dynamem,embodiedvideoagent,worldlines,aem,robomemory}.

SWITCH contributes a complementary evaluation perspective here: its temporally interactive videos of tangible interfaces test state transitions, outcome verification, and recovery after actions~\cite{switchbench}. This motivates checking that an intended transition occurred before superseding a stored belief; SWITCH is a benchmark, rather than a proposed memory architecture.

Many memory systems are naturally append-oriented: new evidence is added while older evidence remains retrievable.
In embodied interaction, this can produce stale beliefs when an object has moved or changed state.
A robust system therefore needs temporal validity and revision, so that historical observations remain available as history while current decisions are conditioned on the latest justified state.

\subsection{Task State: Memory as an Internal State Machine - R3}

\textbf{A diagnostic scenario:} A robot performs a multi-stage household task. Two visually similar scenes can require different next actions because different sub-goals have already been completed, different counts have accumulated, or a previous attempt has failed.

Long-horizon tasks contain latent execution state that may be only weakly expressed in the current visual scene.
Relevant task variables include phase, completed sub-goals, counts, elapsed time since an event, active constraints, and failure history.
The core computational problem is \emph{task-state tracking}: maintaining a procedural state that disambiguates visually similar situations and updates correctly as actions succeed, fail, or need to be retried.
Recent systems make this requirement increasingly explicit. Physical Intelligence's MEM maintains long-term textual notes about task progress together with short-term visual history; MemoryVLA introduces persistent perceptual-cognitive memory for long-horizon manipulation; and NVIDIA's Vesta uses a multimodal memory harness containing historical frames and a running cache of past subtasks~\cite{mem,memoryvla,vesta}. Dedicated benchmarks and policies including RoboMME, RMBench, MIKASA-Robo, RoboMemArena, and KEMO further stress phase ambiguity, counts, ordering, and subtask completion over extended manipulation trajectories~\cite{robomme2026,rmbench2026,mikasa2026,robomemarena2026,kemo}.

Physical time adds a requirement that observation count cannot capture. Temporal Flow Policy retains events while evolving memory over measured intervals, enabling comparisons in which similar observation sequences require different action timing~\cite{tfp2026}. Within R3, a task clock and a completed-action count are therefore different state variables; elapsed duration is also relevant to motion hidden by occlusion.

SimpleARM makes this state-based view explicit: frozen perceptual tools maintain typed relational, reference, progress, and route state for a frozen generalist robot policy~\cite{simplearm2026}.
History is consulted when a sub-goal needs it, and recalled entities are grounded in the current view before execution.

Implicit recurrent tracking can lose sparse variables such as counts or prerequisite order over long trajectories.
Explicit state trackers make those variables easier to inspect but become brittle when tasks branch, actions fail, or the transition logic is uncertain.
The difficult cases are therefore not only remembering the nominal phase, but revising procedural state after execution deviates from the intended plan.

\subsection{Delayed-Relevance Evidence: Memory as Online Selection - R4}

\textbf{A diagnostic scenario:} A robot briefly observes a key on a table and later needs that key to open a cabinet after many unrelated interactions. The useful evidence is short-lived, while its consequence is delayed.

Some observations become relevant long after they disappear from the current context. A transient object location, interaction cue, instruction, or demonstration may influence a decision hundreds of steps later.
This creates an online selection problem because the agent may need to decide what deserves persistence before the future task reveals the true value of the observation.

TaskMem directly formulates memorization as a learnable, task-conditioned policy for deciding what to preserve from streaming multimodal input~\cite{taskmem}. EventVLA predicts the future utility of current visual evidence, KEMO detects task-relevant state-change events, and Worth Remembering uses Bayesian surprise as a task-agnostic gate for episodic storage~\cite{eventvla2026,kemo,worthremembering}. Multi-scale systems such as MEM and long-term multimodal agents such as M3-Agent provide complementary evidence that selective compression and retrieval become increasingly important as interaction history grows~\cite{mem,m3agent}.

WholeBodyMem extends retention to observations preceding goal execution: object events guide keyframe selection for later humanoid loco-manipulation, and its HuMeGo benchmark includes six simulation and six real-robot tasks~\cite{wholebodymem2026}. This highlights pre-goal evidence as a distinct temporal setting. Its object-guided selection should still be distinguished from task-agnostic prospective retention when future relevant entities are unknown.

The core computational problem is \emph{future-utility estimation}. A missed critical event cannot be recovered once its evidence disappears, while storing everything increases memory size, retrieval cost, and interference. The selection criterion is also task-dependent: the same observation can be crucial for one goal and irrelevant for another. This makes selective retention a form of temporal credit assignment conditioned on the task.

\subsection{Belief Maintenance: Conflict Resolution and Active Verification - R5}
\label{sec:r5_belief_maintenance}

\textbf{A diagnostic scenario:} A robot remembers a user's statement that a cup is empty, but subsequently obtains interaction evidence suggesting that it contains liquid. The cup's contents are not visible in the current observation, and the two information sources imply different handling decisions. The robot must determine whether the records concern the same cup and state, assess their reliability, and decide whether further inspection is necessary before acting.

Reliable embodied memory requires an agent to maintain beliefs when retained evidence is incomplete, inconsistent, or no longer valid. A \emph{memory conflict} arises when records or observations support incompatible claims about the same entity, property, and relevant time or task context. Detecting such a conflict first requires aligning these references: an object occupying different locations at different times may simply reflect a valid transition, while two visually similar objects may have been incorrectly associated. R2 emphasizes maintaining current state through action-induced transitions; R5 emphasizes deciding which claims remain justified when observations, action outcomes, or information sources disagree. Its central challenge is therefore \emph{evidence reconciliation}: distinguishing state change from observation error, uncertain execution, mistaken identity, or context mismatch, and maintaining an appropriate degree of confidence in the resulting belief.

Conflict resolution depends on the provenance and reliability of evidence, rather than recency or retrieval similarity alone. A recent textual summary may repeat an earlier mistake, and an action record can describe an intended effect that was never verified. Arkin \etal fuse linguistic, visual, and haptic evidence into a cumulative belief over latent object properties, allowing robots to revise incorrect human-provided information through physical interaction and communicate the disagreement back to the user~\cite{arkin2020multimodal}. This provides a direct embodied example of reconciling conflicting sources. Conflict-Aware Memory addresses a complementary retrieval problem: semantically opposing commands can be close in embedding space, and conflict detection rules filter problematic retrieved contexts~\cite{conflictaware2026}. Its grammatical and interpretation accuracy results support retrieval consistency, while leaving temporal state validity and calibrated physical-state confidence as separate questions. Together, these approaches motivate preserving the source and support of a claim alongside its content.

Belief revision should change the currently supported state while preserving useful historical evidence. DynaMem maintains an online spatial-semantic representation as objects move, appear, or disappear, illustrating the need to synchronize retained scene information with subsequent observations~\cite{dynamem}. Theory of Space identifies a related reasoning failure, \emph{belief inertia}, in which agents fail to overwrite obsolete priors despite contradictory evidence~\cite{tos2026}. These studies motivate a distinction between recording an observation and endorsing it as a current fact. When evidence remains ambiguous, a memory system should be able to retain competing hypotheses or mark a claim as unresolved instead of forcing an unsupported replacement. Conversely, failure to detect an object should be interpreted in relation to visibility and coverage: an occluded object can remain present even when it is absent from the latest image. Revision therefore requires reasoning about what an observation establishes, not merely whether it is newer.

Active verification connects unresolved memory conflicts to the agent's choice of what to observe or ask next. Earlier work by Oh \etal\cite{Oh2016} investigated the joint control of memory, active perception, and action in partially observable Minecraft environments. Their evaluation demonstrates the importance of retaining previously observed information while actively acquiring additional evidence, providing an early example of how memory requirements depend on the interaction between perception and decision-making~\footnote{Though it did not specifically demonstrate conflict resolution between contradictory memory records.}.

Garrett \etal\cite{garrett2020beliefreplan} plan observation and manipulation actions in belief space and re-plan after new evidence, demonstrating how opening containers or removing occluders can resolve uncertainty relevant to execution. GraphEQA uses an incrementally constructed scene graph and visual memory to guide exploration for embodied question answering~\cite{grapheqa}. At the planning level, KnowNo uses conformal prediction to identify uncertainty and request human clarification when needed~\cite{ren2023knowno}. These works support complementary parts of verification and uncertainty-aware action selection; they do not collectively establish a complete solution to memory conflict. For R5, the design implication is that retrieval, re-observation, physical probing, and clarification should be selected according to which disputed variable matters for the next decision and whether the expected benefit justifies the interaction cost.

The core computational problem is \emph{conflict resolution and belief calibration under partial observability}. A capable agent should detect incompatible evidence, identify its scope, revise only the claims whose support has changed, and seek additional evidence when the remaining uncertainty affects action. A convincing evaluation should consequently measure whether the agent accepts valid corrections, resists unreliable contradictions, preserves unrelated correct memories, and changes its behavior appropriately after verification. Relevant failure modes include belief inertia, premature overwriting, inconsistent records, and repeated unnecessary checking. The stronger requirement is to maintain a defensible current belief from historical evidence whose sources, temporal validity, and reliability can differ, while retaining enough provenance to reconsider that belief when new information arrives.

\subsection{Cross-Episode Experience: From Episodes to Knowledge - R6}

\textbf{A diagnostic scenario:} A robot has repeatedly completed the same type of task in many environments. In a new environment, useful experience should reduce search and recovery effort by transferring regularities learned across previous episodes.

Repeated deployment extends historical dependence beyond a single episode. Past trajectories contain environment regularities, interaction patterns, failures, recovery strategies, and reusable skills. The central question is whether the agent merely stores many episodes or extracts knowledge that transfers beyond the specific situations in which it was observed.
This space includes both retrieval-oriented and consolidation-oriented approaches.

Retrieval-Augmented Embodied Agents reuse stored experience; MART learns a multimodal trajectory retriever from interaction-derived preferences and abstracts retrieved trajectories for task-relevant reuse~\cite{raea,mart}. MEMENTO, WorldLines, RoboMemory, and PhysMem explore personalization, stateful reuse, lifelong multi-memory, and abstraction of physical principles~\cite{memento,worldlines,robomemory,physmem}. ASPIRE compounds rollout feedback into persistent skills~\cite{aspire}.

PRACTICE instead makes library maintenance a learned capability: a trained skill learner transforms successful and failed trajectories into edits that add, refine, merge, or remove reusable skills while the executor remains frozen~\cite{practice}. MessyMem addresses complementary environment-specific experience: interaction-derived properties and outcomes augment a persistent spatial scene graph and guide subsequent mobile-manipulation tasks~\cite{messymem2026}.

Recent work distinguishes acquiring reusable experience from choosing its execution context. Robota-Embed learns action-aware matching between the current interaction and recorded continuations~\cite{robotaembed2026}. EmbodiedEvo instead turns physical feedback into code Skills and diagnostic Lessons, then freezes them for fresh agents in unseen scenes~\cite{embodiedevo2026}. Both support R6, but their protocols isolate different steps: selecting a useful past continuation and inheriting consolidated experience.

PAM adds a bounded longitudinal setting: a finite-state assistance memory is revised from daily interaction traces and evaluated over five days, including changed routines~\cite{pam2026}. This supplies more direct evidence of repeated memory revision than a single acquire-then-freeze transfer test, while its scope remains the evaluated assistance routines rather than open-ended lifelong learning.

The core computational problem is \emph{experience abstraction}.
Highly specific memories preserve contextual detail but transfer poorly; aggressive abstraction improves reuse while increasing the risk of applying a rule outside its valid conditions.
MATE shows why retrieval also needs execution compatibility: it transforms retrieved trajectories into task-conditioned execution records through context removal, transition extraction, and verified action normalization~\cite{mate2026}.
MemTransfer evaluates the related transfer problem by changing starting pose, route availability, and historical relevance under a shared frozen policy~\cite{memtransfer2026}.
A large retrievable archive therefore does not by itself imply learning from experience.
The stronger objective is to convert repeated episodes into semantic or procedural knowledge while retaining enough context to know when that knowledge applies.

\subsection{A Computational View of Embodied Memory Requirements}

The six requirements describe distinct reasons why the current observation can be insufficient and distinct computations needed to recover the missing state.
Table~\ref{tab:nonmarkovian_demands} summarizes the taxonomy.

\begin{table}[!htbp]
\centering
\small
\renewcommand{\arraystretch}{1.25}
\setlength{\tabcolsep}{5pt}
\begin{tabularx}{\textwidth}{
    >{\raggedright\arraybackslash}p{0.16\textwidth}
    >{\raggedright\arraybackslash}p{0.19\textwidth}
    Y
    >{\raggedright\arraybackslash}p{0.17\textwidth}
    >{\raggedright\arraybackslash}p{0.16\textwidth}
}
\toprule
\textbf{Memory Requirement} &
\textbf{What Must Persist?} &
\textbf{Core Computational Challenge} &
\textbf{Memory Operation} &
\textbf{Typical Failure} \\
\midrule
Hidden world state &
Entity identity, location, and spatial relations &
Temporal \textbf{entity binding} across occlusion and viewpoint changes &
Retain + retrieve &
Identity drift; forgotten location \\
\addlinespace
Action-induced state change &
Current world state and its temporal versions &
\textbf{State supersession} after the agent changes the environment &
Update + invalidate &
Stale belief \\
\addlinespace
Task state &
Phase, progress, count, constraints, and failures &
\textbf{Latent task-state tracking} across visually similar execution stages &
Accumulate + update &
Repeated action; skipped stage \\
\addlinespace
Delayed-relevance evidence &
Sparse historical cues with future utility &
\textbf{Future-utility estimation} under delayed relevance &
Selective write + recall &
Missed critical event \\
\addlinespace
Active belief maintenance &
World belief and uncertainty &
\textbf{Belief calibration} under uncertainty and environmental change &
Revise + reobserve &
Belief inertia \\
\addlinespace
Cross-episode experience &
Reusable skills, rules, and environmental knowledge &
\textbf{Experience abstraction} across repeated episodes &
Consolidate + transfer &
Overgeneral-\allowbreak ization \\
\bottomrule
\end{tabularx}
\caption{\textbf{A computational taxonomy of non-Markovian embodied memory requirements.}
Each requirement is associated with a distinct latent state, computational challenge, memory operation, and characteristic failure mode.}
\label{tab:nonmarkovian_demands}
\end{table}
\FloatBarrier

The resulting problem-side taxonomy is intentionally independent of any particular memory architecture.
Entity binding, state supersession, task-state tracking, future-utility estimation, belief calibration, and experience abstraction specify what an embodied agent must accomplish when history matters.
Sec.~\ref{sec:memory_architectures} turns to the solution side and asks how existing architectures represent and preserve that history.

The taxonomy suggests a dependency structure that is easy to miss in a paper-by-paper account. Revision requires identifying which entity a new observation concerns; task-progress tracking requires distinguishing an attempted action from its verified effect; and experience abstraction requires knowing which earlier outcome followed from which conditions. Errors can therefore propagate across requirements. An identity swap may look like a state-update failure, while a failed grasp recorded as a completed sub-goal may look like forgetting. Evaluation should localize the earliest incorrect state transition instead of assigning every downstream failure to insufficient retention. This is a diagnostic implication of the taxonomy, not a claim that all implementations must use the same processing pipeline.

A second implication is that the useful lifetime of information differs from its age. The fact that a cup was once on a table can remain relevant to a historical question after becoming invalid for retrieval of the cup. Likewise, a failed action may deserve retention precisely because the corresponding plan should no longer be executed. Systems need to distinguish episodic evidence, current state estimates, and reusable rules, even when these share a physical store. This distinction explains why simply increasing context length can preserve more evidence without improving decisions: the agent must also determine which temporal interpretation of that evidence the present task requires.

\section{Memory Architectures: Representations and Capability Envelopes}
\label{sec:memory_architectures}

Section~\ref{sec:nonmarkovian} described the problem side of embodied memory.
We now examine the solution side: how existing systems represent historical information, how that state is accessed, and which limitations follow from the chosen representation.
We compare \textbf{eight architectural design categories (A1--A8)}: seven recurring mechanism or organization families, followed by hybrid composition as an eighth design category. These categories are intentionally non-exclusive. Fast weights implement a form of recurrence, events may be stored as visual tokens, and experience stores may use symbolic or spatial representations. We separate them when their write, access, or reuse mechanisms introduce a distinct design trade-off. Thus, the grouping is not an ontological claim that eight independent memory substrates exist. Each category has a \emph{capability envelope}: information it makes easy to preserve or access, together with characteristic bottlenecks. The requirements R1--R6 specify what a task needs; A1--A8 specify design choices that may meet those needs.

\subsection{Recurrent State Memory: Compressible Continuous History - A1}

Recurrent memory summarizes interaction history into a continuously updated latent state,
\begin{equation}
m_t=f(m_{t-1},o_t,a_{t-1}).
\label{eq:arch_recurrent}
\end{equation}
ReMem-VLA uses recurrent queries at frame and chunk timescales, while $\mu$VLA augments a VLA with persistent memory tokens trained across timesteps~\cite{rememvla,muvla}.
Gated Memory Policy adds learned gating over historical context~\cite{gmp2026}.
Belief-oriented approaches such as RB-VLA and Action-Effect Memory explicitly compress vision-action history into evolving latent state~\cite{rbvla,aem}.

Recent latent-memory work separates the learning objective from the update dynamics. IMPRINT compresses visual, action, and proprioceptive histories, pretrains memory through history reconstruction and coarse progress prediction, then jointly adapts memory and policy~\cite{imprint2026}. Its progress target is normalized demonstration time, so it supplies temporal supervision without guaranteeing verified sub-goal completion after a failure. Temporal Flow Policy instead makes measured elapsed intervals part of the memory transition~\cite{tfp2026}. These studies identify two orthogonal design questions: which historical content training encourages the state to preserve, and how that state should evolve when observation intervals vary. A compact recurrent memory can address both, but a larger hidden state alone does not specify either.

The main advantage is bounded computation: the persistent state can in principle survive indefinitely without retaining the complete raw trajectory. The corresponding limitation is a fixed-dimensional information bottleneck. Specific historical observations are not directly addressable, and information discarded by the recurrent transition may be unrecoverable. Recurrent designs are therefore efficient for continuously evolving state but weaker when a later decision requires precise access to a particular distant event.

\subsection{Fast-Weight Memory: Historical State in Parameter Space - A2}

Test-time-training architectures store sequence-dependent state in parameters updated during inference.
A generic update takes the form
\begin{equation}
W_t=W_{t-1}-\eta\nabla_W\mathcal{L}_{\mathrm{write}}.
\label{eq:arch_fastweight}
\end{equation}
TTT layers treat fast weights as a recurrent state for sequence modeling~\cite{tttlayer}.

Spatial-TTT uses chunk-wise updates to accumulate global spatial evidence from long visual streams~\cite{spatialttt}; RoboTTT integrates TTT layers into a vision-language-action policy so interaction history is carried through adaptive internal state~\cite{robott}; WAM-TTT places a lightweight adaptive component inside a frozen World-Action Model and updates it from deployment-time evidence~\cite{wamttt}.

T$^2$Mem targets memory-dependent manipulation directly: observation-derived key--value associations are written into fast weights through online self-supervision, while outer action supervision trains their formation and use~\cite{t2mem2026}.
Its memory--policy training alternates the optimized component; at deployment the slow parameters remain fixed and the fast state resets between episodes.
This is within-episode parametric memory, distinct from a library that accumulates across tasks.

Fast weights provide a bounded persistent state whose temporal reach is not tied directly to raw context length.
Their weakness is explicit addressability: the state of a particular historical observation is difficult to recover from parameter space.
They also blur the boundary between memory and adaptation, making the inner-loop objective critical for interpreting what the updated parameters actually encode.

\subsection{Persistent Visual Tokens: Dense Historical Evidence - A3}

Persistent visual memory keeps compressed representations of previous observations rather than collapsing the entire trajectory into a single latent state. NativeMEM and AstraNav-Memory exemplify long-lived visual-token representations~\cite{nativemem,astranavmemory}. These designs retain appearance, pose, motion, and other perceptual evidence that may be lost by textual or symbolic compression.

REKI uses a few action-expert register tokens per frame and caches them across action chunks, avoiding repeated encoding of the corresponding past frames~\cite{reki2026}. This illustrates compact historical access within a generalist VLA rather than a separate retrieval database. The retained context is a recent-history mechanism; control improvements do not alone establish indefinite retention or event-specific random access.

Their main strength is visual fidelity and partial addressability through attention or retrieval over multiple stored tokens.
The cost is that memory can grow with trajectory length unless the system adds compression or eviction.
A second issue is temporal validity: a visual token faithfully records what was observed at some earlier time but does not by itself say whether that evidence is still true.
Dense visual memory therefore solves retention more directly than revision.

\subsection{Event and Keyframe Memory: Sparse Critical Evidence - A4}

Event and keyframe memory assumes that only a small fraction of a long trajectory deserves persistent storage. MemER and SAM2Act+ retrieve informative historical observations, while EventVLA, KEMO, and Worth Remembering explicitly study how useful events should be selected~\cite{memer,sam2act2025,eventvla2026,kemo,worthremembering}. Across these works, selection is driven by notions such as future utility, interaction transition, or surprise.

WholeBodyMem uses object-event-guided keyframes and cross-attention to condition whole-body actions on pre-history~\cite{wholebodymem2026}. Its comparison with uniform memory motivates selecting useful evidence under a budget, while the known object cues constrain how broadly the selection result generalizes.

The advantage is efficiency: storage and retrieval scale with selected events instead of raw trajectory length.
The principal failure mode occurs at write time.
If an observation is not selected when it appears, a later retriever cannot recover it after the evidence has vanished.
Event memories also need a separate mechanism for deciding whether a stored event has become stale.

\subsection{Textual and Symbolic Memory: Semantic Compression - A5}

Textual and symbolic memories compress interaction history into task-relevant semantics such as completed sub-goals, object states, counts, constraints, failures, and plans.
MEM, Mimir, and concept-centric memory approaches instantiate different versions of this representation~\cite{mem,mimir,conceptmemory}.
Mimir, for example, separates task memory from world-oriented state~\cite{mimir}.

SimpleARM illustrates structured task-state memory, retaining typed relations and progress variables rather than unrestricted textual summaries~\cite{simplearm2026}.
PAM uses an explicit finite-state machine with phase-conditioned assistance rules and event-triggered transitions, then revises it from daily traces~\cite{pam2026}. It links procedural state within an interaction to reusable routine knowledge across days, illustrating the many-to-many relation between R3/R6 and one symbolic representation.

Semantic representations are compact, interpretable, and naturally queryable through symbolic keys or language.
They can preserve task progress or high-level state without carrying complete visual history.
The trade-off is perceptual irreversibility: exact geometry, appearance, motion, and subtle object identity can be difficult to reconstruct from a textual summary.
Conflict-Aware Memory introduces rules for detecting conflicting retrieved text and pruning problematic vector-index entries~\cite{conflictaware2026}. Its evidence concerns grammatical and interpretation accuracy in embodied planning, which should be reported separately from physical task completion. TacitVLA makes a complementary distinction between current reasoning and persistent textual history: its action expert does not attend to the newly generated reasoning, while ledger entries are retained for later calls on memory tasks~\cite{tacitvla2026}. Generation can occur during execution, exposing memory-write scheduling as a control-latency design choice. Together, these studies show that semantic memory requires both trustworthy content and an explicit policy for when that content becomes available.

Language therefore provides an efficient abstraction layer when the relevant state is semantic, with weaker coverage of decisions that depend on fine historical visual evidence.

\subsection{Graph and 3D Memory: Spatial Persistence in Dynamic Worlds - A6}

Graph and 3D memories organize historical observations around persistent entities, locations, and spatial relations.
3D-Mem, DynaMem, GraphEQA, and 3DLLM-Mem maintain geometrically grounded representations that support retrieval by location, object, or relation~\cite{3dmem,dynamem,grapheqa,3dllmmem}.
This explicit structure is useful for object permanence, revisitation, navigation, and spatial reasoning across viewpoint changes.

Earlier navigation approaches demonstrated the value of structured spatial memory before the emergence of foundation-model-based embodied agents. Cognitive Mapping and Planning (CMP)~\cite{Gupta2017} combined an incrementally constructed top-down belief map with a differentiable planner, enabling navigation using accumulated observations of previously unseen environments. Semi-Parametric Topological Memory (SPTM)~\cite{Savinov2018} instead represented previously visited locations through a connectivity graph, using learned visual retrieval without requiring an explicit metric map. These approaches illustrate different trade-offs between spatial representation and memory access, showing that effective navigation can depend on preserving actionable spatial structure rather than reconstructing a complete geometric model. 

RANGER combines keyframes, reconstructed geometry, semantic point clouds, and exploration values into a navigational memory map~\cite{ranger}. MessyMem enriches a scene graph with properties learned through interaction and links observations for fine-grained recall~\cite{messymem2026}; Ledger instead records persistent object locations and motion histories for later egocentric spatial queries~\cite{ledger2026}.
These examples distinguish memory for exploration, interaction-derived knowledge, and object-history recall.

MapNav updates a top-down semantic map and adds textual region labels before presenting it to a VLM navigation policy~\cite{mapnav2025}. It demonstrates an interface in which accumulated spatial state replaces historical frames. The resulting abstraction favors spatial and semantic queries over exact recovery of earlier pixels; conclusions about compression should therefore be tied to navigation outcomes and the information discarded by the map.

GeoMem-VLA connects geometry supervision to latent temporal memory~\cite{geomemvla2026}. A geometry teacher supervises compact tokens during training; causal episode memory retains the resulting context for later action generation without running that teacher at inference. This belongs at the boundary of A3, A6, and A8: geometrically informed memory need not be an explicit, renderable map. The design raises a useful comparison with entity stores: compact geometry can improve control while offering less inspectable identity and temporal-validity records.

The central difficulty is keeping a persistent map synchronized with a dynamic world.
Object movement, state changes, pose correction, and data-association errors can leave historical spatial records inconsistent with the current environment.
DynaMem and related systems begin to address dynamic updating, but long-term spatial memory still depends critically on reliable association and temporal validity~\cite{dynamem}.

\subsection{Experience and Skill Memory: Retrieval Versus Consolidation - A7}

Cross-episode memory operates at two common abstraction levels. An early example of experience-based action selection is Neural Episodic Control~\cite{Pritzel2017}, which introduced an external memory of state representations and rapidly updated action-value estimates to support efficient reinforcement learning. Although evaluated outside physical robotics, this approach demonstrated how stored experiences can directly influence subsequent decisions through retrieval, providing an important precedent for more recent episodic-memory architectures.

Episodic approaches such as MemER and OptimusVLA preserve concrete observations or trajectories for later case-based reuse~\cite{memer,optimusvla}.
MART learns which past trajectories are useful to retrieve~\cite{mart}.
Their main challenge is relevance: a semantically similar past episode may differ in physical state, embodiment, or failure condition.
MATE addresses the additional step between retrieval and execution by adapting the retrieved representation to the current task without another LLM call~\cite{mate2026}.

Robota-Embed conditions retrieval on vision, language, and native action, aligning a current-state query with a recorded future continuation~\cite{robotaembed2026}. This makes execution stage part of relevance rather than relying on scene or instruction similarity alone. Its retrieval benchmark and downstream conditioning experiments provide complementary evidence: accurate matching is one component of useful experience, while policy-side integration remains separately evaluated.

A second strategy consolidates repeated experience into reusable concepts, physical principles, or skills.
PhysMem and concept-centric memory move toward this form of abstraction~\cite{physmem,conceptmemory}.
PRACTICE learns reusable skill-library edits from trajectory feedback~\cite{practice}, whereas ProactiveVLA enriches the evidence available for consolidation through proactive environment exploration~\cite{proactivevla2026}.
Consolidation improves transfer and storage efficiency but can lose the contextual conditions under which a learned rule is valid.
MoEC stores subgoal-indexed control experts, grows them after failures, and consolidates redundant experts in Minecraft~\cite{moec2026}. Since its heads are learned online with reinforcement learning, it exemplifies a memory-guided adaptive controller rather than improvement solely through a fixed-policy archive. EmbodiedEvo's code Skills and Lessons provide a contrasting non-parametric reuse interface; its test agents inherit frozen artifacts and do not continue updating the shared store~\cite{embodiedevo2026}. These are different persistence protocols within A7, so neither should be used as automatic evidence for indefinite cross-deployment evolution.

The architectural choice is therefore a trade-off between episodic specificity and reusable abstraction.

\subsection{Hybrid Memory: Combining Multiple Substrates - A8}

Recent systems increasingly combine multiple memory substrates.
MemoryVLA separates working and long-term memory; MEM combines short-term visual evidence with longer-term textual state; MemoryWAM integrates recent observations with compressed longer-lived representations; and Mimir separates world-oriented from task-oriented memory~\cite{memoryvla,mem,memorywam,mimir}.

Industrial ReMEmbR combines clip embeddings, cached visual tokens, and reconstruction latents in a shared spatiotemporal store for robot fleets~\cite{industrialremembr2026}. Expensive scene interpretation can be deferred until a query, shifting cost from routine writes to selected reads. Its evidence concerns retrieval, question answering, traffic mapping, and systems cost; this should be distinguished from demonstrating improved closed-loop manipulation. IMPRINT offers a complementary form of fusion within one latent memory: heterogeneous sensory and action streams can be combined without maintaining multiple independently addressable databases~\cite{imprint2026}. Together, these designs make the level of composition explicit---input streams, retained representations, or distinct stores.

Hybrid designs can allocate different forms of history to representations with different inductive biases. However, architectural composition does not by itself establish functional complementarity. A convincing hybrid design should show under matched conditions that the combined system contributes beyond its individual components and that ablating a component selectively damages the type of information it was intended to preserve. Current evidence for such component-specific complementarity remains limited.

\subsection{Matching Representation to Historical Demand}

Embodied-memory architectures can be viewed as different allocations of representational capacity.
Continuous latent states efficiently compress evolving state; fast weights store adaptive sequence state in parameters; visual memories preserve perceptual evidence; event memories exploit temporal sparsity; symbolic memories preserve semantic task state; spatial memories preserve entity--world relations; and experience memories support reuse across episodes.
Table~\ref{tab:memory_capability} provides a single cross-reference back to the requirements in Sec.~\ref{sec:nonmarkovian}.

\begin{table}[!htbp]
\centering
\small
\renewcommand{\arraystretch}{1.20}
\setlength{\tabcolsep}{5pt}
\begin{tabularx}{\textwidth}{
    >{\raggedright\arraybackslash}p{0.20\textwidth}
    >{\raggedright\arraybackslash}p{0.24\textwidth}
    Y
    Y
}
\toprule
\textbf{Memory Architecture} &
\textbf{Primary Sec.~3 Requirement} &
\textbf{Main Strength} &
\textbf{Main Limitation} \\
\midrule
A1. Recurrent state & Task-state tracking, continuous belief & Fast, bounded-size updates & Information bottleneck; weak random access \\
\addlinespace
A2. Fast-weight / TTT & Continuous belief, spatial or interaction-state accumulation & Bounded adaptive state & Low explicit addressability; memory/adaptation ambiguity \\
\addlinespace
A3. Persistent visual tokens & Entity binding, dense perceptual history & High visual fidelity & Growing context; stale evidence \\
\addlinespace
A4. Event / keyframe & Future-utility estimation & Efficient long-horizon retention & Selection errors can be irreversible \\
\addlinespace
A5. Text / symbolic & Task-state tracking, semantic state & High compression; interpretable & Loss of visual detail \\
\addlinespace
A6. Graph / 3D & Entity binding, spatial belief & Explicit geometric grounding & Update and data-association errors \\
\addlinespace
A7. Experience / skill & R6: case reuse and experience abstraction & Episodic specificity or reusable rules & Mismatch in retrieval; loss of applicability conditions \\
\addlinespace
A8. Hybrid composition & Multiple requirements & Potential cross-scale complementarity & Complexity and attribution difficulty \\
\bottomrule
\end{tabularx}
\caption{\textbf{Capability envelopes of embodied-memory architectures.}
The requirement column is a diagnostic cross-reference to Sec.~\ref{sec:nonmarkovian}; A1--A7 are non-exclusive mechanism or organization families, while A8 denotes their composition. Episodic retrieval and skill consolidation are two modes within A7, not additional top-level categories.}
\label{tab:memory_capability}
\end{table}
\FloatBarrier

\subsection{Design Implications Across Categories}

\textbf{The first design decision is which information loss the task can tolerate.} Recurrent states and fast weights exchange direct access to historical observations for compact updates; visual stores retain detail at an increasing access cost; symbolic summaries trade perceptual fidelity for semantic precision. The appropriate comparison therefore fixes the historical query as well as the memory budget. A counter and a visually ambiguous object identity can require the same retention duration but very different representations. A useful design should preserve the smallest state sufficient for its intended decisions, while retaining supporting evidence when that state may need correction. This principle explains why one representation can perform well on procedural tracking and poorly on fine-grained recall without either result establishing a universal ranking.

\textbf{Memory failures occur at different stages and require different remedies.} Selection determines whether evidence enters memory; representation determines which details survive; retrieval determines which retained evidence reaches the policy; revision determines which information remains valid; and policy integration determines whether available evidence changes action. Increasing retrieval capacity cannot recover an event discarded at write time. Increasing storage cannot correct a policy that attends to an obsolete entry. A practical diagnostic progression is to supply the missed evidence, then supply the correct retrieved entry, and finally supply the correct current state, measuring which intervention repairs the behavior. These interventions are proposed analytical controls; they should not be confused with results already demonstrated by the reviewed systems.

\textbf{Hybrid memory needs an agreement mechanism as well as complementary representations.} A visual frame, a task note, and a scene graph can refer to the same object while disagreeing about its current location or whether an action succeeded. Without provenance and update rules, adding stores can increase the number of mutually inconsistent beliefs. A useful hybrid should specify which observation or verified action can supersede each record, how an entity is linked across stores, and how uncertainty triggers re-observation. Component ablations should consequently test conflict resolution as well as recall: retaining an accurate old frame is compatible with making an incorrect present-state decision. The resulting design objective is coordinated maintenance of decision-relevant state under a bounded resource budget.

\textbf{Cross-episode improvement requires an explicit persistence and transfer protocol.} A7 describes an organization for reusable experience, but its name does not establish that any particular experiment retains state between episodes. Every method should report what is reset, what persists, whether the base policy changes, and which test conditions differ from the stored cases. Retrieval can improve a new episode through a close match without learning a reusable rule; consolidation can produce a concise rule that fails when its applicability conditions change. Evidence for sustained improvement therefore requires held-out situations, conflicting or obsolete experience, and comparison with a frozen archive under matched interaction budgets. This separates benefits of a growing memory from benefits of repeated exposure or continued policy training.

\section{Benchmark Coverage and Diagnostic Validity}
\label{sec:benchmarks}

Long-horizon difficulty, partial observability, and memory dependence characterize different aspects of an embodied task.
A trajectory may span hundreds of actions while remaining locally solvable whenever the current observation reveals all information required for the next decision.

A much shorter task can impose stronger memory requirements when a decisive observation disappears before its consequences are resolved. As embodied evaluation expands from tabletop manipulation to mobile manipulation, active perception, long-horizon planning, and household activities, this distinction becomes increasingly important.

We therefore examine benchmarks along two complementary axes. The first is the \emph{control level} at which historical information is consumed: high-level planning, hierarchical whole-stack execution, low-level visuomotor control, or dedicated memory evaluation. The second is \emph{memory diagnosticity}: how directly benchmark success can be attributed to historical information that the agent itself must preserve, update, and later use.

We use \emph{memory diagnosticity} to describe how directly a benchmark isolates behavior that depends on information no longer available in the current observation, rather than on trajectory length or general task difficulty.

\subsection{Evaluation Protocols Across Control Levels}
\label{sec:benchmark_protocols}

\textbf{High-level evaluations score the use of observations to choose actions or answers, with much of motor execution supplied by the interface.} EmbodiedBench~\cite{embodiedbench2025} presents visual observations and instructions, executes the model's selected actions, and reports task completion across its environment and capability subsets. Its high-level household tasks primarily test grounding and sub-goal choice through semantic actions, whereas its lower-level environments impose a different action burden. ESI-Bench~\cite{esibench2026} evaluates final-answer accuracy after observation gathering, comparing a single view, active multi-view inputs, active exploration, and oracle-trajectory views. This comparison helps distinguish obtaining useful evidence from reasoning over supplied evidence, although both can depend on retained history. EMem-Bench~\cite{emembench2026} supplies prior interaction histories before subsequent embodied execution; Sec.~\ref{sec:case_emem} examines its protocol. These interfaces make history-conditioned planning or reasoning accessible to study, while success remains sensitive to perception, action selection, and the information provided externally.

\textbf{Hierarchical evaluations must expose how much progress information is supplied to the planner.} RoboCerebra~\cite{robocerebra2025} couples a high-level planner to a lower-level policy and reports state-transition completion, exact plan agreement, plan efficiency, and action-completion interpretation. Its state-transition measure credits satisfied key conditions, so it should not be conflated with binary full-episode success; its anchor-based planner comparison also supplies transition structure. BEHAVIOR-1K~\cite{behavior1k} evaluates physically executed household goals whose satisfaction depends on object states and relations. For memory analysis, the important protocol variables are the permitted observations and the source of task-state information: simulator predicates may legitimately score completion while remaining unavailable to the policy. Task success under onboard sensing has strong deployment relevance but combines navigation, control, grounding, and historical state. Authors should identify the observation regime and distinguish evaluator access to ground truth from policy access before treating either suite as evidence of memory.

\textbf{Low-level evaluations measure whether remembered information can be converted into successful robot motion.} Language-conditioned suites such as LIBERO, RoboCasa, RoboTwin~2.0, and VLABench commonly report rollout success under task-specific goal predicates~\cite{libero2023,robocasa2024,robotwin2025,vlabench2025}. Such scores jointly depend on choosing the correct target and realizing the required physical interaction. CALVIN~\cite{calvin2022} adds instruction chains: completion rates for successive subtasks and average chain length measure reliable composition, while the evaluator's successive instructions supply part of procedural state. RoboCasa365~\cite{robocasa365} additionally distinguishes multi-task, foundation-model, and lifelong-learning settings; sequential-training retention measures a different process from persistence of observations within a deployment episode. Dedicated memory subsets make the history variable clearer, but a failed grasp can still mask correct recall. RoboDojo~\cite{robodojo2026} is particularly useful here because its \textit{Memory} and \textit{Long-Horizon} splits are separate and its partial-progress score complements binary success; Sec.~\ref{sec:case_robodojo} examines the \textit{Memory} protocol.

\subsection{What the Scores Measure and What They Establish}
\label{sec:benchmark_scores}

\textbf{A benchmark metric has meaning only together with its observation interface and completion rule.} Table~\ref{tab:mainstream_benchmarks} makes these elements explicit for representative general benchmarks. The evaluation target column describes the competence jointly tested by the rollout, whereas the last column states what remains unresolved about historical dependence. In particular, a goal-predicate success rate, a chain-completion measure, an answer-accuracy score, and a partial-progress score should retain their distinct meanings even when each is reported as a percentage. Their aggregation also matters: weighting tasks equally and weighting capability families equally can produce different overall rankings. Memory comparisons should report the relevant task family rather than relying only on an overall score dominated by other capabilities.

\begin{table*}[!t]
\centering
\scriptsize
\renewcommand{\arraystretch}{1.18}
\setlength{\tabcolsep}{3.5pt}
\begin{tabularx}{\textwidth}{>{\raggedright\arraybackslash}p{0.15\textwidth}YYYY}
\toprule
\textbf{Benchmark} & \textbf{Policy interface} & \textbf{Reported outcome} & \textbf{What is jointly evaluated} & \textbf{Memory interpretation limit} \\
\midrule
LIBERO~\cite{libero2023} & Language, visual/proprioceptive input; robot actions & Goal-completion success; sequential-learning studies also examine transfer and retention & Language grounding, manipulation, and knowledge transfer & Training-time retention and within-episode history use require separate evidence \\
\addlinespace
CALVIN~\cite{calvin2022} & Successive language subtasks and sensor inputs; continuous actions & Success across 1--5 subtasks; average completed chain length & Compositional execution and generalization & Successive instructions supply part of task-progress state \\
\addlinespace
RoboCasa~\cite{robocasa2024} & Language and robot observations; manipulation rollout & Task success under kitchen goal conditions & Generalist manipulation across tasks and environments & A visible goal state may reduce the need for distant recall \\
\addlinespace
RoboCasa365~\cite{robocasa365} & Multi-task/foundation-model rollouts;  & Task success; performance across sequential training phases & Atomic/composite generalization and learned competence retention & Continued optimization is distinct from deployment memory \\
\addlinespace
RoboTwin~2.0~\cite{robotwin2025} & Language and multimodal robot observations; bimanual actions & Rollout task success under randomized conditions & Coordinated manipulation and domain robustness & Randomization alone does not isolate historical dependence \\
\addlinespace
VLABench~\cite{vlabench2025} & Language-conditioned robot control; separate VLM/workflow evaluations & Execution success; protocol-specific reasoning/plan metrics & Grounding, compositional reasoning, and execution & VLM answer/plan quality and physical completion are different outcomes \\
\addlinespace
RoboDojo~\cite{robodojo2026} & Standardized policy observation/action loop & Binary success and partial-progress score by capability & Five capabilities, including separate Memory and Long-Horizon splits & Memory score includes recall, grounding, and motor execution \\
\addlinespace
Embodied-\allowbreak Bench~\cite{embodiedbench2025} & Vision and instructions; environment-specific semantic/atomic actions & Task success by environment and capability subset & High-level and low-level planning, reasoning, navigation, and manipulation & Action abstraction changes which state the policy must maintain \\
\addlinespace
EMem-Bench~\cite{emembench2026} & Supplied multimodal history and a new task; semantic actions & Terminal-state success and Error Recurrence Rate, averaged across four families & History-conditioned task execution & Observed objects, updated locations, failed interactions, and recurring corrections \\
\addlinespace
ESI-Bench~\cite{esibench2026} & Passive observations or actively selected sensing/interaction; final answer & Answer accuracy across observation-access regimes & Evidence acquisition and spatial reasoning & Correct answers do not alone isolate persistent-memory use \\
\addlinespace
RoboCerebra~\cite{robocerebra2025} & Planner sub-goals coupled to an execution policy & Key-state completion, plan match, plan efficiency, completion QA & Decomposition, monitoring, reflection, and execution & State-transition credit and anchor scaffolding affect interpretation \\
\addlinespace
BEHAVIOR-1K~\cite{behavior1k} & Household robot rollouts; observation privileges depend on protocol & Activity completion against goal conditions & Whole-stack household competence & End-to-end failure does not identify a memory error \\
\bottomrule
\end{tabularx}
\caption{\textbf{Evaluation interfaces, outcomes, and memory implications of general embodied benchmarks.} Entries summarize the named benchmark protocols rather than prescribing one universal sensor or action configuration. A score should be interpreted using its own completion rule and observation privileges.}
\label{tab:mainstream_benchmarks}
\end{table*}
\FloatBarrier

\subsection{Two Cases of Memory Evaluation: High-Level and Low-Level}
\label{sec:benchmark_cases}

The following two cases connect an actual task to its evidence, execution, and scoring protocol. EMem-Bench tests historical-state use through high-level embodied actions; RoboDojo tests history-conditioned manipulation through robot control. Each case identifies the retained variable and the behavior it should change, while the interpretation distinguishes the measured outcome from a proposed additional diagnostic.

\subsubsection{High-Level Case: Updated Object Location in EMem-Bench}
\label{sec:case_emem}

EMem-Bench~\cite{emembench2026} evaluates the use of supplied history during later physical interaction, rather than recall through a verbal answer alone. Its 2,554 episodes provide ordered multimodal history, a new task, and an executable action space. Four task families cover incidental visual evidence, changing world state, interaction-revealed constraints, and transfer of recurring corrections. Actions are executed in simulation; success requires the target terminal state, and the overall score equally averages family success rates. Error Recurrence Rate additionally counts repeated memory-related errors relative to decision opportunities. In the paper's Dynamic Tracking example (Appendix C.8), a remote control moves from a drawer to an armchair before the target execution. Correct behavior retrieves it from the armchair, requiring the earlier location to be superseded (R2) while preserving object identity (R1). A return to the drawer is a stale-state choice; a failed pickup at the armchair can instead be an execution failure. The supplied history controls evidence availability but leaves autonomous acquisition and long-term continual learning outside this test.

This case makes temporal validity observable at the first location-dependent decision. Our proposed diagnostic is to pair plausible histories that end with different remote locations while matching the current view and task, then record the first chosen destination as well as terminal success. If that choice follows the latest supported location, the policy demonstrates use of updated historical state even when later execution fails. If the policy searches both locations indiscriminately and eventually succeeds, terminal success alone provides weaker evidence about revision. The benchmark's recurrence measure is useful for this distinction, but it should be read alongside the actual action trace: it measures specific repeated errors under its event definitions rather than general memory confidence. These interpretations separate retention, supersession, and successful use without requiring a new architectural category.

\subsubsection{Low-Level Case: Conveyor Matching in RoboDojo}
\label{sec:case_robodojo}

RoboDojo's \textit{Memory} split contains six tasks and is scored separately from its eight \textit{Long-Horizon} tasks~\cite{robodojo2026}. In \texttt{match\_and\_pick\_from\_conveyor}, an earlier object disappears and the policy must later pick a candidate of the same category. The retained variable is the reference category (R1), and the delayed candidate makes earlier evidence useful after it is absent (R4). Its standard simulation protocol uses 50 episodes per task; model experiments generally repeat evaluation across training seeds. It reports binary completion and a partial-progress score, then averages within capability splits and equally across the five splits overall. These outcomes test memory-conditioned manipulation, including visual discrimination and timely physical capture, rather than isolated category recall. The task is valuable because simply extending a workflow cannot resolve the missing reference. The distinct \texttt{imitate\_sorting\_sequence} task stresses remembered demonstration order, illustrating that the \textit{Memory} split spans several historical variables rather than one retention mechanism.

The useful failure split is whether the robot selected the wrong category or failed to grasp a correctly selected moving object. Our proposed analysis records target selection before contact and then reports capture and final completion conditional on correct selection. Swapping the earlier reference while keeping later candidate observations comparable tests whether selection follows history; cue removal and a same-backbone context control test how much retention is needed. This would strengthen attribution beyond the published aggregate score. Partial progress can reveal where execution stopped, but its task-specific rubric is not a standalone memory-accuracy measure. Finally, this case concerns the simulation \textit{Memory} split: RoboDojo's separate real-world suite is a complementary deployment evaluation, not a set of matched physical replications of all six memory tasks. Keeping these interfaces explicit prevents a simulation recall result from being interpreted as complete real-world memory validation.

\subsection{Memory-Specific Coverage and Remaining Gaps}
\label{sec:benchmark_gaps}

Dedicated suites complement the two cases by controlling other historical variables. RMBench, RoboMME, MIKASA-Robo, RoboMemArena, LIBERO-Mem, MemoryBench, RoboTwin-MeM, MemMimic, ReMemBench, MemoryEQA, and LMEE introduce counting, occlusion, historical reference, delayed evidence, or exploration requirements~\cite{rmbench2026,robomme2026,mikasa2026,robomemarena2026,liberomem2025,sam2act2025,eventvla2026,gmp2026,prism2026,memoryeqa2025,lmee2026}. MemTransfer instead varies starting pose, route availability, and historical relevance under a shared policy, helping assess reuse under changed conditions~\cite{memtransfer2026}. These protocol differences motivate the conservative requirement mapping in Table~\ref{tab:benchmark_coverage}: task coverage and indirect pressure should remain distinct from causal isolation of a memory computation.

\begin{table*}[!t]
\centering
\scriptsize
\renewcommand{\arraystretch}{1.16}
\setlength{\tabcolsep}{5.0pt}
\begin{tabularx}{\textwidth}{
>{\raggedright\arraybackslash}p{0.25\textwidth}
C C C C C C}
\toprule
\textbf{Benchmark} &
\textbf{EB} &
\textbf{SS} &
\textbf{TS} &
\textbf{FU} &
\textbf{BC} &
\textbf{EA} \\
\midrule

RMBench~\cite{rmbench2026}
& \checkmark & & \checkmark & & & \\

RoboMME~\cite{robomme2026}
& \checkmark & \checkmark & \checkmark & & & \\

LIBERO-Mem~\cite{liberomem2025}
& \checkmark & & & & & \\

RoboMemArena~\cite{robomemarena2026}
& \checkmark & \checkmark & \checkmark & & & \\

MemoryBench~\cite{sam2act2025}
& \checkmark & & & & & \\

RoboTwin-MeM~\cite{eventvla2026}
& & & & $\triangle$ & & \\

MemMimic~\cite{gmp2026}
& & \checkmark & & & & $\triangle$ \\

ReMemBench~\cite{prism2026}
& \checkmark & & & & & \\

MIKASA-Robo~\cite{mikasa2026}
& \checkmark & \checkmark & \checkmark & & & \\

MT-HM3D~\cite{memoryeqa2025}
& \checkmark & & & & & \\

LMEE~\cite{lmee2026}
& & & & $\triangle$ & $\triangle$ & \\

HuMeGo~\cite{wholebodymem2026}
& \checkmark & $\triangle$ & & $\triangle$ & & \\

EMem-Bench~\cite{emembench2026}
& \checkmark & \checkmark & $\triangle$ & $\triangle$ & $\triangle$ & \checkmark \\

MemTransfer~\cite{memtransfer2026}
& & $\triangle$ & & & $\triangle$ & $\triangle$ \\
\bottomrule
\end{tabularx}
\caption{\textbf{Coverage of the six memory requirements from Sec.~3 in memory-oriented benchmarks.}
EB (R1): Entity Binding; SS (R2): State Supersession; TS (R3): Task-State Tracking;
FU (R4): Future-Utility Estimation; BC (R5): Belief Calibration; EA (R6): Experience Abstraction. These are the computational operations associated with the same six requirements, not six additional memory types.
\checkmark denotes systematic task coverage, while $\triangle$ indicates partial or indirect pressure without dedicated evaluation.
The mapping is our interpretation of task protocols, not a claim of causal isolation; HuMeGo concerns pre-goal object history, with selection and state-change coverage treated as indirect; EMem-Bench's EA mark concerns its controlled generalization family, while MemTransfer probes reuse under mismatch rather than learning an abstraction mechanism.}
\label{tab:benchmark_coverage}
\end{table*}
\FloatBarrier

HuMeGo adds a whole-body setting in which observations preceding goal execution must support later actions~\cite{wholebodymem2026}. We mark its object retention directly and state-change and selection demands conservatively: an object-guided keyframe protocol is informative about pre-history use without isolating open-ended prospective selection.

The strongest coverage currently lies in entity binding, state supersession, and task-state tracking.
RoboMME, RoboMemArena, MIKASA-Robo, RMBench, and related benchmarks repeatedly test whether an observation, object state, interaction, or procedure remains available after the corresponding evidence disappears.
This provides increasingly strong tests of episodic retention.

In many of these tasks, the historical variable that deserves preservation is identifiable early.
A demonstration specifies the relevant procedure, an occlusion event identifies the object to track, or the instruction explicitly indicates what should later be recalled.
The resulting challenge is primarily to preserve known-relevant information over time.

Future-utility estimation moves the decision to an earlier stage.
When an observation first appears, its eventual importance may still be unknown.
RoboTwin-MeM begins to expose this problem through task-critical transient events~\cite{eventvla2026}.
ESI-Bench approaches a related selection problem from the perception side by asking agents to acquire observations that reduce uncertainty~\cite{esibench2026}.
These two operations should be distinguished: information \emph{acquisition} asks what the agent should observe, while information \emph{retention} asks what among those observations should remain available. Current benchmarks provide stronger supervision for information acquisition and passive retention than for deciding prospectively which acquired observations deserve persistence.

Belief calibration is also beginning to emerge as an evaluation pressure. ESI-Bench reports cases in which agents maintain an early hypothesis despite later contradictory observations~\cite{esibench2026}. LMEE similarly connects episodic memory with proactive exploration~\cite{lmee2026}. These settings probe whether historical information remains reliable after new evidence arrives. Dedicated metrics for memory confidence, staleness, and revision remain limited.

Experience abstraction has begun to receive targeted evaluation, but remains less comprehensively tested than within-episode retention. Most memory benchmarks operate within one episode and ask the agent to retain specific events or states. MemMimic provides initial cross-trial pressure~\cite{gmp2026}, while broader lifelong-learning benchmarks such as RoboCasa365 evaluate retention across sequential training phases~\cite{robocasa365}. EMem-Bench supplies controlled regularity-transfer tasks, and MemTransfer examines reuse when execution conditions differ~\cite{emembench2026,memtransfer2026}.

These are useful initial tests, but neither alone establishes sustained improvement over an open-ended sequence of self-collected episodes. A stronger test of embodied experience abstraction would keep the base policy fixed and ask whether an evolving non-parametric memory can extract reusable regularities across episodes and improve later behavior.

The benchmark audit reveals a clear imbalance in current coverage. Entity binding, state supersession, and task-state tracking receive comparatively direct evaluation in RoboMME, RoboMemArena, MIKASA-Robo, RMBench, and related suites.
Future-utility estimation appears in early form through transient-event and information-selection settings, while belief calibration is only indirectly probed through contradictory evidence, active sensing, or exploration.

Cross-episode experience abstraction remains less thoroughly isolated in benchmark suites: EMem-Bench and MemTransfer add controlled generalization and transfer tests, but sustained consolidation over self-collected episodes remains underexplored, while broader lifelong-learning suites often conflate memory with continued parameter learning.

Method-level longitudinal evaluations are also becoming more concrete: PAM~\cite{pam2026} revises assistance memory across days, while EmbodiedEvo~\cite{embodiedevo2026} freezes acquired experience for independent test agents. These protocols answer different questions about revision and transfer. Benchmark coverage should therefore record whether evaluation continues to update shared memory, rather than labeling both settings as equivalent continual learning.

A second gap concerns diagnostic attribution.
General-purpose benchmarks can create substantial memory pressure while still mixing memory with perception, planning, manipulation, generalization, and recovery.
Even dedicated memory benchmarks can permit current-state leakage or provide scaffolding that externalizes part of the historical state. Accordingly, Section~\ref{sec:evaluation} turns from describing existing benchmarks to the separate question of what evidence is needed to establish genuine memory use.

\section{Towards Requirement-Grounded Evaluation of Embodied Memory}
\label{sec:evaluation}

Evaluation should begin by establishing that a particular decision requires historical information. The paired-state criterion in Eq.~\eqref{eq:history_dependence} provides a constructive test: keep the goal and current observation comparable, vary a plausible preceding history, and require a different correct action. Current-observation, bounded-context, full-history, and persistent-memory policies should then be compared under matched backbones, observation privileges, and execution interfaces. Full history provides an access reference but still requires successful reasoning, while a memory-augmented policy may benefit from extra computation or training. This makes task construction and experimental controls equally necessary. The two cases in Sec.~\ref{sec:benchmark_cases} illustrate the distinction: EMem-Bench exposes whether a policy chooses a location justified by updated history, and RoboDojo exposes whether a remembered reference controls physical target selection. Reporting these decisions alongside terminal success helps determine whether historical information reached the behavior that needed it.

Diagnostic interventions should match the historical variable and distinguish failure stages. Entity-history swaps test R1; newly contradicted locations test R2; visually matched phases with different completed counts test R3; a query revealed only after a capacity-limited write phase tests R4; varying evidence reliability with optional costly re-observation tests R5; and held-out situations with exceptions to a learned rule test R6. For prospective selection, explicitly announcing the future target changes the problem to retention of known-relevant evidence. Likewise, conflicting timestamps alone do not measure calibrated uncertainty: confidence and information-seeking must respond to reliability and environmental change. CanTest offers a related capability-level approach, using generated tests and separate perception, memory, planning, and decision oracles for VLN~\cite{cantest2026}. Its memory oracle compares explicit memory descriptions with historical visual annotations, so it supplies an attribution tool while leaving latent memory and the action-level necessity of every recalled detail partly unresolved. Combining intermediate-state inspection with plausible history interventions can localize selection, retrieval, revision, and execution errors more convincingly than either technique alone.

Memory quality should be reported jointly with its temporal reach and deployment cost. Performance curves over cue-to-use delay, distractor load, object count, and state changes reveal which information is lost as demands increase. Measured duration should also be varied independently of frame count, as motivated by Temporal Flow Policy~\cite{tfp2026}. Storage, retrieval latency, update latency, policy latency, and interaction cost reveal the resources used to preserve it. TacitVLA's concurrent generation of entries for future ledger use illustrates why write scheduling matters alongside representational capacity~\cite{tacitvla2026}. A representation that preserves a useful cue but delays control can be unsuitable for moving-object capture, whereas the same latency may be acceptable in household planning. Hybrid systems additionally need component-specific ablations and conflict tests to establish whether different stores contribute complementary evidence and maintain consistent state. Proposed interventions should also preserve plausible histories wherever possible: arbitrary shuffling can create malformed inputs and introduce a distribution shift unrelated to forgetting. These controls connect claims about efficiency and specialization to the actual historical variable under test.

Cross-episode evaluation should specify what persists and demonstrate transfer beyond the acquired cases. Each protocol must state which memory resets, whether the policy or expert heads are updated, how acquisition and evaluation scenes differ, and whether new test experience enters the shared store. Robota-Embed's stage-sensitive continuation retrieval and EmbodiedEvo's frozen inherited Skills and Lessons illustrate distinct ways for experience to influence later execution~\cite{robotaembed2026,embodiedevo2026}. Their relevance to sustained improvement depends on the persistence protocol, not the label attached to the system. Useful controls include a frozen archive, an expanding unprocessed archive, matched acquisition budgets, and tests with conflicting or obsolete experience. Transfer should improve held-out decisions while limiting regression on earlier conditions. Across temporal scales, this yields three research priorities: selecting useful evidence before its eventual query is known, preserving history while revising current beliefs, and consolidating reusable knowledge with its applicability conditions. Progress on these priorities can turn memory from an additional input channel into a dependable state-maintenance capability for embodied deployment.

\section{Final Remarks}
\label{sec:conclusion}

Embodied memory supports action by preserving decision-relevant information after its originating evidence leaves the current observation. This survey organized that role through six overlapping requirements: hidden world state, action-induced state change, task state, delayed-relevance evidence, active belief maintenance, and cross-episode experience. A separate architectural axis compared eight design categories, comprising seven recurring mechanism or organization families and hybrid composition. Their relationship is many-to-many: a requirement can be served by different representations, and one implementation can support several requirements. Making this distinction explicit connects architectural choices to the information that a task actually needs, while avoiding an ambiguous enumeration of memory types.

The comparison yields three principal conclusions. First, retention, access, and temporal validity are separate sources of success and failure: an agent can preserve a historical observation perfectly and still act incorrectly because that observation is obsolete or retrieved in the wrong context. Second, representation should be selected according to tolerable information loss and required access, with compact latent state, visual detail, symbolic structure, spatial grounding, and reusable experience serving different decision demands. Hybrid composition adds value when these capabilities are coordinated and experimentally attributable. Third, benchmark scores support claims only within their observation and execution protocols. The EMem-Bench and RoboDojo cases expose different uses of history: selecting an action from updated world state and converting a remembered visual reference into robot control. Their scoring protocols also show why aggregate success combines memory with perception, planning, and execution.

The field's next challenge is to make retained experience dependable under changing tasks and environments. This requires selecting useful evidence before all future queries are known, revising beliefs when actions or observations invalidate them, and transferring accumulated experience without losing its applicability conditions. The literature and benchmark coverage reviewed here remain uneven, and the proposed framework is a diagnostic synthesis rather than an exhaustive or mutually exclusive taxonomy. Nevertheless, it offers a common basis for stronger comparisons: identify the missing historical variable, specify how the system maintains it, and demonstrate under controlled interventions that it improves the later decision. Progress on these criteria would make memory a more reliable foundation for embodied agents that operate beyond isolated episodes.

\bibliographystyle{unsrtnat}
\bibliography{ref}

\end{document}